\documentclass[runningheads]{llncs}
\usepackage{soul, color, xcolor}
\usepackage{graphicx}
\usepackage{CJKutf8}
\usepackage{CJK}
\usepackage{graphicx}
\usepackage{float}
\usepackage{hyperref}
\hypersetup{hypertex=true,
            colorlinks=true,
            linkcolor=blue,
            anchorcolor=blue,
            citecolor=blue}
\usepackage{booktabs}
\usepackage{multirow}
\usepackage{biblatex}
\usepackage{bm}
\usepackage{amssymb}
\usepackage{xspace}
\usepackage{enumitem}
\usepackage{amsmath}
\usepackage{bbding}

\newcommand{\ie}{\textit{i.e.,}\xspace}

\newcommand{\eg}{\textit{e.g.,}\xspace}

\newcommand{\ignore}[1]{}

\begin{document}
%
\title{Towards Effective Ancient Chinese Translation: Dataset, Model, and Evaluation}
\titlerunning{Towards Effective Ancient Chinese Translation}
%
\author{Geyang Guo$^1$ \and
Jiarong Yang$^1$ \and
Fengyuan Lu$^1$ \and
Jiaxin Qin$^1$ \and
\\Tianyi Tang$^1$ \and
Wayne Xin Zhao$^{1,2}$\thanks{Corresponding author.}}
\authorrunning{G. Guo et al.}
%
\institute{$^1$Gaoling School of Artificial Intelligence, Renmin University of China\\
$^2$Beijing Key Laboratory of Big Data Management and Analysis Methods\\
\email{\{guogeyang,yangjiarong001,lufengyuan,2020201476\}@ruc.edu.cn,\\steventianyitang@outlook.com,batmanfly@gmail.com}}
\maketitle              
\begin{abstract}
Interpreting ancient Chinese has been the key to comprehending vast Chinese literature, tradition, and civilization. In this paper, we propose \textbf{Erya} for ancient Chinese translation. 
From a dataset perspective, we collect, clean, and classify ancient Chinese materials from various sources, forming the most extensive ancient Chinese resource to date. 
From a model perspective, we devise Erya training method oriented towards ancient Chinese. We design two jointly-working tasks: disyllabic aligned substitution (DAS) and dual masked language model (DMLM). 
From an evaluation perspective, we build a benchmark to judge ancient Chinese translation quality in different scenarios and evaluate the ancient Chinese translation capacities of various existing models.
Our model exhibits remarkable zero-shot performance across five
domains, with over +12.0 BLEU against GPT-3.5 models and better human evaluation results than ERNIE Bot. Subsequent fine-tuning further shows the superior transfer capability of Erya model with +6.2 BLEU gain. 
We release all the above-mentioned resources at \href{https://github.com/RUCAIBox/Erya}{https://github.com/RUCAIBox/Erya}.

\end{abstract}

\section{Introduction}







Ancient Chinese literature is a long-cherished cultural legacy not only for Chinese people but for all of humanity. However, due to the evolution of the Chinese language over time, it can be challenging for modern readers to fully comprehend these works. In order to bring ancient Chinese literature back to modern life, we conduct our work towards effective ancient Chinese translation.

Currently, the translation of ancient Chinese is predominantly carried out by professionals, but this process is time-consuming and labor-intensive, impeding the widespread dissemination and comprehension of the knowledge embedded in ancient Chinese.
With the help of deep learning, recent work has focused on using the pre-training strategy in machine translation for ancient Chinese~\cite{DBLP:journals/corr/abs-2009-11473, DBLP:conf/nlpcc/YangCC21, DBLP:conf/acl-lchange/ChangSYD21}. However, these works merely follow the paradigm of English-centered pre-training, ignoring the characteristics of ancient Chinese.

In this work, we present \textbf{Erya} targeting effective ancient Chinese translation.
We first collect and clean ancient Chinese corpus to construct the \textit{Erya dataset} including both monolingual ancient data and ancient-modern parallel data, and further classify them according to textual and chronological characteristics. With the dataset and classification criteria, we propose \textit{Erya benchmark} for comprehensive ancient Chinese translation evaluation.

Considering that recent large language models (LLMs)~\cite{llm_survey} can be improved by supervised fine-tuning~\cite{DBLP:conf/nips/Ouyang0JAWMZASR22}, we utilize the parallel data from Erya dataset to train our model. 
In order to incorporate the characteristics of ancient Chinese, 
we devise two training strategies for Erya model in analogy to the process of ancient Chinese learning. 
Firstly, we propose \textit{disyllabic aligned substitution (DAS)} to narrow the representation gap between aligned ancient-modern word pairs.
In addition, we design \textit{dual masked language model (DMLM)} with a bidirectional decoder to optimize both ancient and modern representations. 



The contributions of our work are listed as follows:  
\begin{enumerate}
\item We build \textbf{Erya dataset}, the largest ancient Chinese dataset to the best of our knowledge, and design \textbf{Erya benchmark} for comprehensive evaluation.
\item We propose the \textbf{Erya model}, which is specifically designed for ancient Chinese translation and leverage the features of ancient Chinese.
\item We evaluate the performance of some large language models and commercial translation services on Erya benchmark, and the results validate the superiority of Erya model on both zero-shot and fine-tuning settings.
\end{enumerate}

\section{Related Work}
\subsection{Pre-training in Neural Machine Translation}
Leveraging large scale corpora, pre-training is an effective method to acquire general language representations and achieve superior performance on various downstream tasks. Current works such as GPT-3~\cite{DBLP:conf/nips/BrownMRSKDNSSAA20} and BERT~\cite{DBLP:conf/naacl/DevlinCLT19} are based on the Transformer architecture, leading to improvements in language understanding tasks. Moreover, numerous sequence-to-sequence pre-trained models (\eg BART~\cite{DBLP:conf/acl/LewisLGGMLSZ20} and T5~\cite{DBLP:journals/jmlr/RaffelSRLNMZLL20}) aim to correspond better with text generation tasks.

Furthermore, researchers transfer the idea of pre-training from monolingual to multilingual. For example, XLM~\cite{DBLP:conf/nips/ConneauL19}, mBART~\cite{DBLP:journals/tacl/LiuGGLEGLZ20}, and mT5~\cite{DBLP:conf/naacl/XueCRKASBR21} utilize multilingual corpora to learn representations of multiple languages for translation. In addition, several works~\cite{DBLP:conf/emnlp/ChiDMHSMHSW21,DBLP:journals/corr/abs-2106-13736,DBLP:conf/emnlp/LinPWQFZL20} pre-train models on language parallel pairs to narrow the semantic distance among the same word across different languages. CeMAT~\cite{DBLP:conf/acl/LiLZWL22} further proposes a new pre-training task, which combines both multilingual corpora and parallel data, to achieve a better translation performance.


\subsection{Ancient Chinese Domain Tasks}
In the realm of ancient Chinese information processing, fundamental tasks consist of automated sentence segmentation, word segmentation, word representation, dataset construction, and ancient-modern translation.

Relevant studies on sentence segmentation~\cite{cheng-etal-2020-integration} and word segmentation~\cite{TP-toolbox-web} are proposed to deal with the issue that ancient Chinese corpora lack punctuation. In addition, since the existence of polysemous words, some studies focus on ancient Chinese word representation~\cite{shu-etal-2021-gu}. As for the construction of corresponding datasets, there are studies focusing on the alignment approach~\cite{DBLP:journals/talip/LiuYQL20}, polysemous words~\cite{DBLP:conf/naacl/PanWOK22}, and poetry understanding~\cite{Leiliu}.

For ancient-modern translation, AnchiBERT~\cite{DBLP:journals/corr/abs-2009-11473} proposed the first pre-trained language model in the ancient Chinese domain with the masked token prediction task, while Guwen-UNILM~\cite{DBLP:conf/nlpcc/YangCC21} applied a two-stage pre-training framework using ancient Chinese corpora and ancient-modern translation pairs. Besides, a semi-supervised translation model~\cite{DBLP:conf/acl-lchange/ChangSYD21} was developed to predict both the translation and its particular era, improving the performance with additional chronological context. However, these studies merely apply the general translation method without considering the unique characteristics of ancient Chinese.



\section{Erya Dataset}
We construct an open-source collection, named \textbf{Erya dataset}, consisting of ancient monolingual corpus and ancient-modern parallel corpus. Our Erya dataset currently stands as the most extensive ancient Chinese resource.


\subsection{Data Collection and Cleaning}                       
In order to construct our raw corpus, we first crawl data from the Internet~\cite{wyw5156} and collect open source data~\cite{TP-toolbox-web2,TP-toolbox-web3, TP-toolbox-web1,TP-toolbox-web4,TP-toolbox-web5,TP-toolbox-web6}. Most of the ancient Chinese texts are written throughout the dynasties of ancient China spanning from 1000 BC to AD 1600. Then, we carry out a cleaning process to eliminate noise from the raw corpus and ensure consistent formatting as follows:
\begin{itemize}[leftmargin=*]
\itemsep0em 
\item \textbf{Noise Filtering}
We design three rules to filter out noises: (1) delete tokens other than Chinese characters (\eg Arabic numerals and English words); (2) simplify traditional Chinese characters; and (3) unify non-text characters such as punctuation (\eg converting \begin{CJK*}{UTF8}{gbsn}「\end{CJK*} to \begin{CJK*}{UTF8}{gbsn}“\end{CJK*}).

\item \textbf{De-duplication}
Since our raw corpus is collected from different sources, \ie the contents may overlap with each other, we further leverage MinHash algorithm for efficient de-duplication. We compare text pieces from various sources and only keep one copy if the similarity between the two pieces is below 0.5.




\item \textbf{Automatic Punctuation}
Since some collected monolingual texts lack punctuation, we employ the guwen-punc~\cite{TP-toolbox-web} toolkit to add punctuation.


\end{itemize}

\subsection{Data Classification}
After data processing to construct a cleaned Erya dataset for ancient Chinese, we further propose a method to classify it based on textual characteristics. We incorporate the traditional ``Four-Branch Classification''~\cite{sibu} and a chronological classification~\cite{RN02, four} to consider the characteristics of different dynasties, grammar, sentence structure, and background knowledge. The details of the whole classification system are as follows:



\begin{itemize}
    \item \textbf{History}: It includes historical literature that document dynastic changes, significant events and notable figures.
    \begin{itemize}
        \item \textbf{Old Chinese}: Pre-Qin to Han Dynasty (before AD3)
        \item \textbf{Middle Chinese}: Three kingdoms to Song Dynasty (AD4 to AD12)
        \item \textbf{Early Modern Chinese}: Yuan to Qing Dynasty (AD13 to AD19)
    \end{itemize}
    \item \textbf{Article}: It contains various literary works, including poetry, prose, philosophy works and literary criticism. 
    \item \textbf{Novel}: It mixes ancient and quasi-modern styles, thus having its own unique textual genres and linguistic features.
\end{itemize}



\subsection{Statistics}

In summary, our Erya dataset consists of a total of 88,808,928 ancient Chinese sentences and 1,941,396,399 characters with an average sentence length of 21.9. The parallel corpus within it comprises a total of 2,087,804 sentences and 84,769,383 characters. The average sentence length of ancient and modern sentences in the parallel part are 17.3 and 23.3 characters respectively. We present a comparison between existing ancient Chinese resources in Table~\ref{tab:dataset}. We can see that our Erya dataset stands out as the most abundant resource among both monolingual and parallel data at present.



\begin{table}[htbp]
\small
\caption{\textbf{Comparison between existing ancient Chinese resources and Erya dataset. Mono. and Para. denote the number of characters in monolingual corpus and parallel corpus (including source and target), respectively.}}
\label{tab:dataset}
\centering
\resizebox{1.0\linewidth}{!}{
\begin{tabular}{c|ccccccc|c}
\toprule
\textbf{Datasets}&Time~\cite{DBLP:conf/acl-lchange/ChangSYD21}&Guwen~\cite{DBLP:conf/nlpcc/YangCC21}&Anchi~\cite{DBLP:journals/corr/abs-2009-11473}&DBU~\cite{TP-toolbox-web1}&Daizhige~\cite{TP-toolbox-web2}&THUCC~\cite{TP-toolbox-web3}&NLD~\cite{DBLP:journals/talip/LiuYQL20}&\textbf{Erya dataset}\\
\midrule
\textbf{Mono.}&4M&1,743M&395M&89M&1,743M&-&-&\textbf{1,941M}\\
\midrule
\textbf{Para.}&1.6M&2.5M&31.5M&56.5M&-&13.4M&37.9M&\textbf{84.8M}\\
\bottomrule
\end{tabular}
}
\end{table}

The monolingual ancient texts can be utilized to learn the general knowledge in ancient Chinese, while the parallel data can bridge the linguistic gap between ancient and modern Chinese. Furthermore, considering the scarcity of a benchmark for ancient Chinese translation, we design \textbf{Erya benchmark}, a subset of the parallel data based on our classification criteria. We have taken into account the diverse textual characteristics of ancient Chinese literature and linguistic evolution throughout the ages. The detailed statistics are listed in Table~\ref{tab:stat}.


\begin{table}[htbp]
\caption{\textbf{Statistics of Erya benchmark. \#ASL is the average sentence length.}}
\small
\label{tab:stat}
\centering
\resizebox{0.8\linewidth}{!}{
\begin{tabular}{c|ccc|c|c}
\toprule
&&\textbf{History}&&\multirow{2}{*}{\textbf{Article}}&\multirow{2}{*}{\textbf{Novel}} \\
&\textbf{Old}&\textbf{Middle}&\textbf{Early Modern}& \\
\midrule
\multirow{2}{*}{\textbf{Title}}&\textit{Book of Han}&\textit{New Tang History}&\textit{Ming History}&\textit{Xu Xiake's Travels}&\textit{Taiping Guangji}\\
&(\begin{CJK*}{UTF8}{gbsn}汉书\end{CJK*})&(\begin{CJK*}{UTF8}{gbsn}新唐书\end{CJK*})&(\begin{CJK*}{UTF8}{gbsn}明史\end{CJK*})&(\begin{CJK*}{UTF8}{gbsn}徐霞客游记\end{CJK*})&(\begin{CJK*}{UTF8}{gbsn}太平广记\end{CJK*})\\
\midrule
\textbf{\#Train}&18,646&9,396&66,730&16,649&45,162 \\
\midrule
\textbf{\#Valid}&2,331&1,174&8,341&2,081&5,645 \\
\midrule
\textbf{\#Test}&2,331&1,175&8,342&2,082&5,646 \\
\midrule
\textbf{\#ASL}&21.2&20.5&21.5&25.1&20.0 \\
\bottomrule
\end{tabular}
}
\end{table}

\section{Erya Model}
\label{sec:model}

\begin{figure}[htbp]
  \centering
  \includegraphics[width=0.95\textwidth]{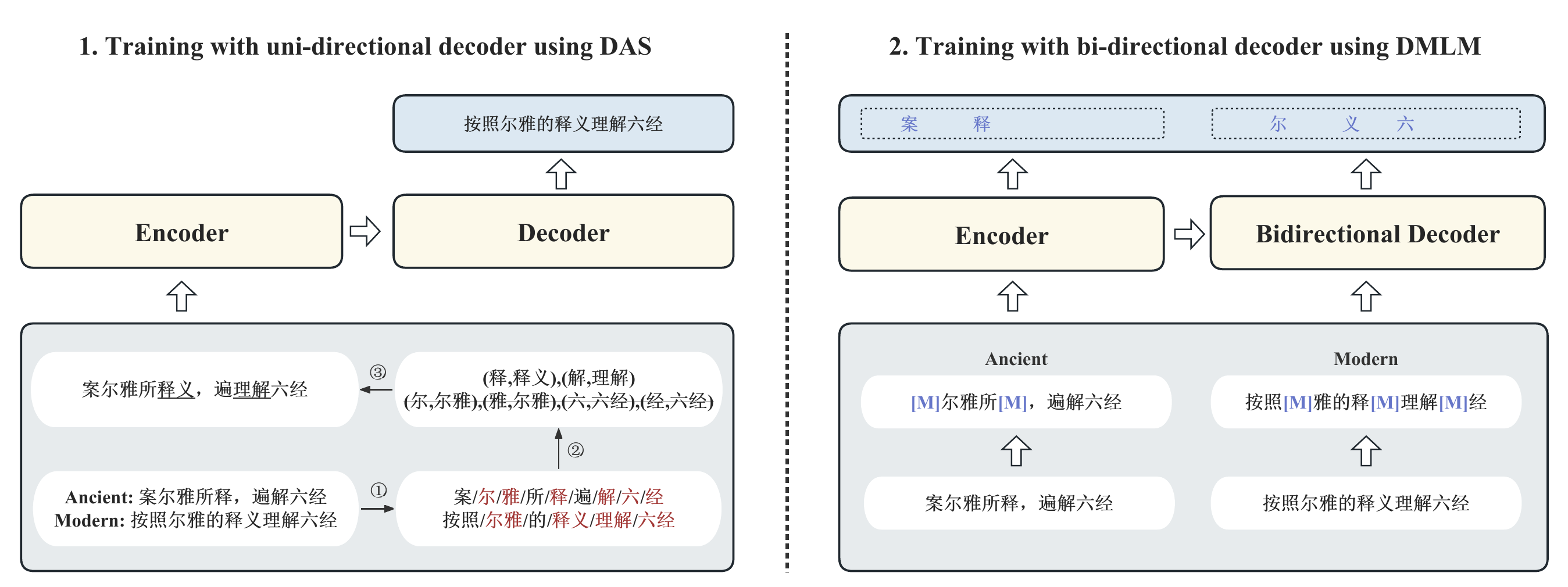}
  \caption{\textbf{The illustrated training framework of Erya model.}}
  \label{fig:example}
\end{figure}

In this section, we propose our Erya model which is specially designed for ancient Chinese translation. Considering the recent success of supervised fine-tuning~\cite{DBLP:conf/nips/Ouyang0JAWMZASR22}, our Erya model is further fine-tuned on the existing CPT~\cite{DBLP:journals/corr/abs-2109-05729} model using Erya parallel data (\ie ancient-modern translation pairs) with two training tasks: \textbf{disyllabic aligned substitution (DAS)} and \textbf{dual masked language modeling (DMLM)}. 
The overall framework is illustrated in Figure~\ref{fig:example}. 


Formally, we denote parallel data as $\mathcal{D}$, with each translation pair as $(X, Y)$.

\subsection{Disyllabic Aligned Substitution}
In order to boost the linguistic similarity between ancient and modern Chinese, we devise a substitution approach to close the word representation gap between the two languages following previous work~\cite{DBLP:conf/emnlp/LinPWQFZL20}.


\subsubsection{Training Objective:}
The disyllabic aligned substitution is a generation task: given a noised ancient sentence where the ancient-modern aligned words are randomly substituted by its modern translation, the model needs to generate the translation autoregressively using the following training objective:
\begin{equation}\label{eq:DAS}
    \mathcal{L}^{DAS} = - \sum_{(X, Y)\in \mathcal{D}} \log P (Y|\tilde{X}) = - \sum_{(X, Y)\in \mathcal{D}} \log P (y_1, \dots, y_i |\tilde{X}, y_{<i}),
\end{equation}
where $\tilde{X}=C(X, Y)$ is the disyllabic word alignment function.


\subsubsection{Disyllabic Word Alignment:}\label{Disyllabic Align Strategy}
In ancient Chinese translation, disyllabic expansion~\cite{caomiao} is a widespread method, \ie a monosyllabic ancient character is often translated into a disyllabic word (\eg \begin{CJK*}{UTF8}{gbsn}解\end{CJK*} $\rightarrow$ \begin{CJK*}{UTF8}{gbsn}理解\end{CJK*} in Figure
~\ref{fig:example}). We explore our parallel dataset and find that 99.8\% of the pairs contain such alignment, and an average of 62.5\% of characters in source sentences can be aligned.

We design an efficient and effective strategy called disyllabic word alignment in three steps:
(1) Source sentence $X$ is split at the character level and target sentence $Y$ is segmented by the THULAC toolkit\footnote{\url{http://thulac.thunlp.org/}}.
(2) A monosyllabic character of the source and a disyllabic word of the target are matched if they contain a common character, forming a set of alignment pairs. Then, we filter out the target words that occur in the source sentence, which may be proper nouns and do not need to be translated. 
(3) We replace the aligned characters in $X$ with its counterpart in $Y$ by probability $P_{DA}=0.7$ (experimented in Section~\ref{sec:pda}). 
 



\subsection{Dual Masked Language Modeling}
\label{sec:dmlm}
As suggested by previous work~\cite{DBLP:conf/acl/LiLZWL22}, applying masked language modeling for both the encoder and decoder can enhance the model's representational ability and further improve translation performance. With such insight, we train our model using dual masked language modeling (DMLM). 

We randomly mask tokens at both sides independently with a dynamic probability of $[0.1, 0.2]$ and $[0.2, 0.5]$ for ancient and modern sentences, respectively. Each token has an 80\% probability of being masked, a 10\% probability of being replaced with a random token, and a 10\% probability of being unchanged following BERT~\cite{DBLP:conf/naacl/DevlinCLT19}. As for the pair $(X, Y)$, we denote the masked-token subset as $(X^{mask}, Y^{mask})$. Hence, the DMLM objective can be formulated as:

\begin{equation}\label{eq:dmlm}
\begin{aligned}
\small
\mathcal{L}^{DMLM} & = &\lambda& \mathcal{L}^{enc} &+& (1-\lambda) \mathcal{L}^{dec} \\
& = - &\lambda& \sum_{x_i\in X^{mask}} \log{P(x_i | \hat{X})} &-& (1-\lambda) \sum_{y_i \in Y^{mask}} \log{P(y_i | \hat{X}, \hat{Y})},
\end{aligned}
\end{equation}
where $\hat{X}$ and $\hat{Y}$ denote the unmasked tokens in $X$ and $Y$, respectively. We set the coefficient $\lambda=0.3$ following~\cite{DBLP:conf/acl/LiLZWL22}. Note that the decoder is bi-directional and it predicts the masked tokens non-autoregressively.

\subsection{Erya Multi-task Training}
\label{sec:mul}
Combining the above objectives~\ref{eq:DAS} and~\ref{eq:dmlm}, the final training loss is:
\begin{equation} \label{eq:combine}
    \mathcal{L} = (1-\mu) \mathcal{L}^{DAS} + \mu \mathcal{L}^{DMLM},
\end{equation}
where $\mu$ is a weight to balance two objectives. In Section~\ref{sec:lambda}, we find $\mu=0.3$ can achieve a better translation performance. In order to mitigate the gap between training and inference, we include an additional epoch of translation training directly using parallel data.

\section{Experiment}
\subsection{Experimental Setup}
In our experiments, we utilize Erya benchmark for zero-shot and fine-tuning evaluation using automatic and human evaluations. 


\subsubsection{Baseline Methods}
In order to better evaluate Erya model, we choose several baselines for comparison: 
(1) CPT-base~\cite{DBLP:conf/naacl/DevlinCLT19} has demonstrated satisfactory results on Chinese tasks. 
(2) AnchiBERT~\cite{DBLP:journals/corr/abs-2009-11473} and Guwen-UNILM~\cite{DBLP:conf/nlpcc/YangCC21} are translation models designed for ancient Chinese.
(3) text-davinci-003, gpt-3.5-turbo\footnote{\url{https://platform.openai.com/docs/models/gpt-3-5}}, and ERNIE Bot (\begin{CJK*}{UTF8}{gbsn}文心一言\end{CJK*})\footnote{ \url{https://yiyan.baidu.com/}} are recent LLMs and can achieve excellent performance on various downstream tasks. text-davinci-003 and gpt-3.5-turbo are from the GPT-3.5 series and are mainly trained on English corpus, while ERNIE Bot specifically considers Chinese.
(4) Baidu Translate\footnote{\url{https://fanyi-api.baidu.com/product/11}} is a commercial translation service that supports ancient Chinese translation.


\subsubsection{Implementation Details}
(1) For CPT-base and Guwen-UNILM, we fine-tune them  following the recommended hyper-parameters from the original papers. 
(2) For AnchiBERT, we combine it with a randomly-initialized decoder and follow the further training practices and hyper-parameters in its original paper. 
(3) For text-davinci-003 and gpt-3.5-turbo, we leverage OpenAI API and prompt the system with the sentence ``\begin{CJK*}{UTF8}{gbsn}将这句话翻译为现代汉语：\end{CJK*}".
(4) For Baidu Translate, we just input the ancient Chinese and receive the translated output.
(5) For ERNIE Bot, we can only randomly select 100 sentences using manual input for human evaluation due to the lack of an API usage license. 

For our Erya model training, we apply the AdamW optimizer~\cite{DBLP:conf/iclr/LoshchilovH19} with a learning rate of 3e-5 and a batch size of 256. We call the model after multi-task training (Equation~\ref{eq:combine}) \textbf{Erya4FT} and the model with an additional translation training \textbf{Erya}.
Then, we utilize Erya for zero-shot translation and employ Erya4FT to fine-tune on specific datasets with a batch size of 64. We employ the text generation library TextBox~\cite{tang-etal-2022-textbox} for implementation.


\subsubsection{Evaluation Metrics}

We use BLEU~\cite{DBLP:conf/acl/PapineniRWZ02} and BERTScore~\cite{DBLP:conf/iclr/ZhangKWWA20} for automatic evaluation: (1) BLEU applies N-gram matching to measure the similarity between candidate output and reference translation. (2) BERTScore calculates the similarity between each token in candidate and reference sentences by utilizing contextual embeddings. 
\subsection{Experiment Results}
\begin{table}[h]
    \centering
    \label{table3}
    \caption{\textbf{Translation performance comparison on Erya benchmark. 
    Bold and underlined fonts indicate the best and the second best score.}}
    \label{tab:main result}
    \resizebox{1.0\columnwidth}{!}{
    \small
    \begin{tabular}{c|c|cc|cc|cc|cc|cc|cc}
    \hline
    \multirow{2}{*}{\textbf{Settings}}&\multirow{2}{*}{\textbf{Model}}& \multicolumn{2}{c|}{\textit{\textbf{Book of Han}}} &\multicolumn{2}{c|}{\textit{\textbf{New Tang History}}}&\multicolumn{2}{c|}{\textit{\textbf{Ming History}}}&\multicolumn{2}{c|}{\textit{\textbf{Taiping Guangji}}}&\multicolumn{2}{c|}{\textit{\textbf{Xu Xiake's Travels}}}&\multicolumn{2}{c}{\textbf{Average}}\\
    &&BLEU&BERTScore&BLEU&BERTScore&BLEU&BERTScore&BLEU&BERTScore&BLEU&BERTScore&BLEU&BERTScore\\\hline
    \multirow{6}{*}{\textbf{Zero-shot}}
    &\textbf{AnchiBERT} &\underline{24.4}&81.1&\underline{30.9}&84.0&30.8&83.5&\underline{19.0}&78.2&19.3&80.5&24.9&81.5  \\
    &\textbf{Guwen-UNILM} &22.3&\underline{81.3}&30.1&\underline{84.9}&31.8&84.1&18.6&\underline{78.5}&20.8&81.2&24.7&\underline{82.0}  \\
    &\textbf{text-davinci-003} &15.2&74.9&20.6&77.2&20.7&76.7&14.9&75.2&18.4&79.1&18.0&76.6\\
    &\textbf{gpt-3.5-turbo} &17.7&76.4&20.5&76.4&21.5&76.6&16.8&76.1&20.1&80.3&19.3&77.2\\
    &\textbf{Baidu Translate} &24.3&79.7&29.1&82.4&\textbf{38.9}&\underline{84.5}&16.9&74.9&\textbf{43.5}&\underline{86.8}&\underline{30.5}&81.7\\
    &\textbf{Erya} &\textbf{29.9}&\textbf{85.3}&\textbf{34.5}&\textbf{88.4}&\underline{37.1}&\textbf{88.0}&\textbf{24.1}&\textbf{81.6}&\underline{34.2}&\textbf{88.2}&\textbf{32.0}&\textbf{86.3} \\
    \hline
    \multirow{3}{*}{\textbf{Fine-Tuning}}
    &\textbf{Guwen-UNILM}&28.7&83.0&38.9&87.1&36.6&85.0&22.2&79.8&35.2&86.2&32.3&84.2\\
    &\textbf{CPT} &\underline{32.8}&\underline{84.3}&\underline{41.3}&\underline{88.1}&\underline{43.7}&\underline{87.8}&\underline{25.8}&\underline{81.2}&\underline{41.3}&\underline{88.1}&\underline{37.0}&\underline{85.9} \\
    &\textbf{Erya4FT} &\textbf{34.9}&\textbf{85.2}&\textbf{42.4}&\textbf{88.5}&\textbf{44.3}&\textbf{88.0}&\textbf{27.1}&\textbf{81.7}&\textbf{42.5}&\textbf{88.4}&\textbf{38.2}&\textbf{86.4} \\
    \hline
    \end{tabular}
    }
    \label{tab:my_label}
\end{table}
\subsubsection{Zero-shot Translation}
From the result in Table~\ref{tab:main result}, we can obverse that Erya can achieve superior ancient Chinese translation ability among all baselines with the best BLEU and BERTScore. Erya can outperform previous ancient Chinese models (+7 BLEU) and GPT series models (+12 BLEU) by a large margin. When compared with commercial Baidu Translate, we also achieve better BERTScore (+4.6) and BLEU (+1.5) scores. 

Note that Baidu Translate may have been trained on the examples on our Erya benchmark, while we exclude the benchmark data during the Erya training. The BERTScore of Erya in the zero-shot can even reach the performance of Erya4FT fine-tuned on specific datasets.



\subsubsection{Fine-tuned Translation}
To further evaluate the domain transfer capability of Erya, we fine-tune Erya4FT on specific datasets. 
After fine-tuning, Erya4FT obtains +6.2 BLEU gain than zero-shot and is +1.2 BLEU superior to the fine-tuned CPT on average. This demonstrates Erya's potential of accustoming and transferring to a specific domain and the effectiveness of our training task. 

In summary, our model is rather small (145M) but effective compared with GPT series (175B). Large language models may struggle to effectively process specialized corpus content, such as ancient Chinese. This further emphasizes the importance of designing smaller models specifically tailored for ancient Chinese.
Hence, Erya is a more parameter-efficient and well-performing option for effective ancient Chinese translation on both zero-shot and fine-tuning scenario. 

\subsection{Further Analysis}
\subsubsection{Ablation Study}

To analyze the effect of the proposed Erya multi-task training in Equation~\ref{eq:combine}, we design the following variants to compare the performance of zero-shot and fine-tuned translation: 

\begin{itemize}[leftmargin=*]
\itemsep0em 
\item[$\bullet$] w/o DAS: the variant removes DAS in the training stage.
\item[$\bullet$] w/o DMLM: the variant does not use DMLM when training.
\item[$\bullet$] w/o dynamic mask: we remove the dynamic mask mentioned in Section~\ref{sec:dmlm}, but with a fixed ratio of 0.15 and 0.35 for the encoder and decoder respectively.
\item[$\bullet$] w/o translation training: we remove the additional translation training after the DAS and DMLM training.

\end{itemize}


\begin{table}[htbp]
\caption{\textbf{Ablation analysis on Erya benchmark using the BLEU metric.}}
\label{tab:ablation}
\centering
\small
\resizebox{1.0\columnwidth}{!}{
\begin{tabular}{c|c|c|c|c|c|c|c}
\toprule
&\textbf{Models}&\textbf{\textit{Book of Han}}&\textbf{\textit{New Tang History}}&\textbf{\textit{Ming History}}&\textbf{\textit{Xu Xiake's Travels}}&\textbf{\textit{Taiping Guangji}}&\textbf{Avg.}\\
\midrule
\multirow{4}{*}{\textbf{Zero-shot}}&\textbf{Erya}&\textbf{29.9}&\textbf{34.5}&\textbf{37.1}&\textbf{34.2}&\textbf{24.1}&\textbf{32.0} \\
&\textbf{w/o DAS}&28.9&33.3&36.8&32.3&23.4&30.9 \\
&\textbf{w/o DMLM}&28.7&32.9&36.6&32.0&23.2&30.7 \\
&\textbf{w/o translation training}&27.6&33.1&35.5&31.3&22.4&30.0 \\
\midrule
\multirow{4}{*}{\textbf{Fine-Tuning}}&\textbf{Erya4FT}&\textbf{34.9}&\textbf{42.4}&\textbf{44.3}&\textbf{42.5}&\textbf{27.1}&\textbf{38.2} \\
&\textbf{w/o DAS}&34.4&41.7&44.0&42.0&26.5&37.7 \\
&\textbf{w/o DMLM}&33.9&41.9&44.1&42.1&26.8&37.8 \\
&\textbf{w/o dynamic mask}&34.4&42.2&43.9&42.2&26.9&37.9 \\
\bottomrule
\end{tabular}}
\end{table}

From Table~\ref{tab:ablation}, we can see that: 
(1) Both DAS and DMLM consistently have a positive effect across five evaluation datasets. For the zero-shot setting, Erya achieves +1.1 BLEU with DAS and +1.3 BLEU with DMLM. 
For the fine-tuning setting, Erya4FT achieves +0.5 BLEU with DAS and +0.4 BLEU with DMLM. 
(2) Dynamic mask ratio plays a positive role in model performance. Enabling it results in +0.3 BLEU gain for the fine-tuning scenario. 
(3) The additional translation is beneficial in the zero-shot translation scenario, which can alleviate the gap between downstream translation and denoising training.
(4) The positive effect of each component is more significant in zero-shot than in fine-tuning, which implies our strategies are more vital for zero-shot translation. 


\subsubsection{Performance Comparison w.r.t. Substitution Ratio} \label{sec:pda}
\begin{figure}[htbp]
    \centering
    \includegraphics[width=0.9\linewidth]{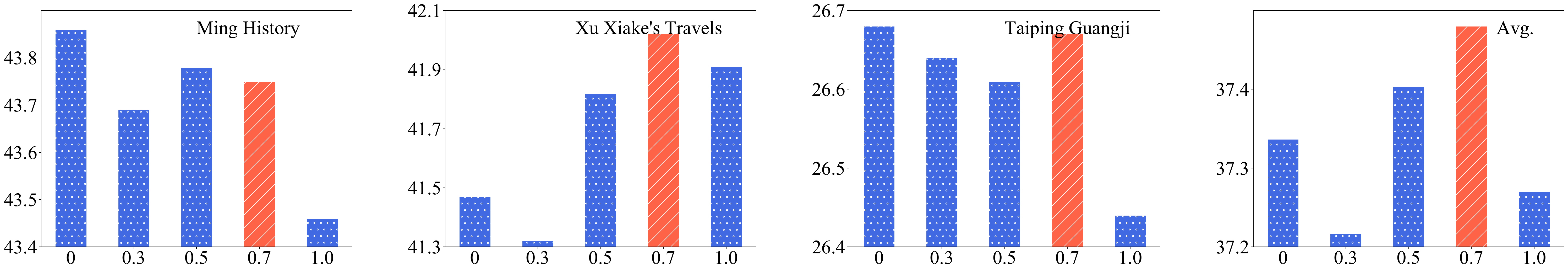}
    \caption{\textbf{BLEU scores of different substitution ratios.}}
    \label{fig:ana1}
\end{figure}

The substitution ratio $P_{DA}$ in DAS affects the degree to close the representations of ancient and modern Chinese. 
Here, we vary the ratio to study its influence on translation. We select three sets of the Erya benchmark, covering all the literary styles. From the results in Figure~\ref{fig:ana1}, we set $P_{DA}$ as 0.7 for better overall performance.



\subsubsection{Performance Comparison w.r.t. the Weights of two Losses} \label{sec:lambda}
The weight $\mu$ in Equation~\ref{eq:combine} balances the model's performance toward representation and generation. Here we vary $\mu$ among: 0.3, 0.5, and a decreasing strategy from 0.5 to 0.
Table~\ref{tab:ana2} shows that with $\mu=0.3$  the model performs better almost in each dataset. We speculate that the two tasks can incorporate better by emphasizing more on generation capacity.


\begin{table}[htbp]
\caption{\textbf{BLEU scores with different $\mu$.}}
\label{tab:ana2}
\centering
\small
\resizebox{1.0\columnwidth}{!}{
\begin{tabular}{c|c|c|c|c|c|c}
\toprule
\textbf{$\mu$}&\textbf{\textit{Book of Han}}&\textbf{\textit{New Tang History}}&\textbf{\textit{Ming History}}&\textbf{\textit{Xu Xiake's Travels}}&\textbf{\textit{Taiping Guangji}}&\textbf{Avg.}\\
\midrule
\textbf{0.3}&\textbf{34.9}&42.4&\textbf{44.3}&\textbf{42.5}&\textbf{27.1}&\textbf{38.2}\\
\textbf{0.5}&34.2&\textbf{42.5}&43.8&42.0&26.8&37.9 \\
\textbf{0.5$\rightarrow$0}&\textbf{34.9}&42.4&44.1&42.1&26.7&38.0\\
\bottomrule
\end{tabular}
}
\end{table}



\subsection{Human Evaluation}
In addition to automatic evaluation, we further carry out a human evaluation. We randomly select 20 ancient texts from five categories of Erya benchmark, and gather the translations produced by gpt-3.5-turbo, ERNIE Bot, Baidu Translate, CPT, Erya, and Erya4FT. These translations are randomized to facilitate impartial human assessment.

We invite three domain experts majored in Ancient Chinese Literature evaluators to assess the quality of generated texts based on three criteria: faithfulness (\begin{CJK*}{UTF8}{gbsn}信\end{CJK*}, degree of accuracy exhibited by the translated text concerning the ancient text), expressiveness (\begin{CJK*}{UTF8}{gbsn}达\end{CJK*}, degree of fluency and clarity of the text), and elegance (\begin{CJK*}{UTF8}{gbsn}雅\end{CJK*}, degree of appropriateness and elegance of the text). In addition, we design the overall criterion to evaluate how likely the generated text is produced by human.
We adopt a 5-point Likert scale as the scoring mechanism, in which 5-point means ``very satisfying'', and 1-point means ``very terrible''.

From the results shown in Table~\ref{tab:human}, we can see that our Erya model outperforms almost all the baselines by a significant margin under both zero-shot and fine-tuning settings. This demonstrates the effectiveness of our multi-task training approach and our high-quality datasets.

\begin{table}[htbp]
\caption{\textbf{Human evaluation on Erya benchmark.}}
\label{tab:human}
\centering
\small

\resizebox{0.7\columnwidth}{!}{
\begin{tabular}{c|c|c|ccc}
\toprule
\textbf{Settings}&\textbf{Models}&\textbf{Overall}&\textbf{Faithful}&\textbf{Expressive}&\textbf{Elegant}\\
\midrule
\multirow{4}{*}{\textbf{Zero-shot}}&\textbf{Baidu}&2.83&3.23&2.80&2.75\\
&\textbf{gpt-3.5-turbo}&\textbf{3.55}&3.42&3.80&\textbf{3.75} \\
&\textbf{ERNIE Bot}&3.45&3.48&3.75&3.48 \\
&\textbf{Erya}&3.47&\textbf{4.22}&\textbf{3.81}&3.46\\
\midrule
\multirow{2}{*}{\textbf{Fine-tuning}}&\textbf{CPT}&3.72&3.93&3.84&3.54 \\
&\textbf{Erya4FT}&\textbf{3.80}&\textbf{4.01}&\textbf{3.92}&\textbf{3.61} \\
\midrule
~&\textbf{Gold}&4.23&4.27&4.40&4.05\\
\bottomrule
\end{tabular}
}
\end{table}


\section{Conclusion}
In this paper, we introduce Erya for ancient Chinese translation consisting of Erya dataset, model, and benchmark.
Erya dataset is currently the largest ancient Chinese corpora collection including both monolingual corpus and ancient-modern parallel data. 
We further propose a multi-task learning combining DAS and DMLM 
 to train Erya model. 
Finally, we conduct comprehensive evaluation using Erya benchmark. Extensive experiments have validated the superior capability of Erya model under both zero-shot and fine-tuning settings. 


\section*{Acknowledgments}
This work was partially supported by National Natural Science Foundation of China under Grant No. 62222215, Beijing Natural Science Foundation under Grant No. 4222027, and Beijing Outstanding Young Scientist Program under Grant No. BJJWZYJH012019100020098. Xin Zhao is the corresponding author. Special thanks to Manman Wang for the advice on ancient Chinese.

\printbibliography

\end{document}